\documentclass[conference]{IEEEtran}
\IEEEoverridecommandlockouts
\usepackage{cite}
\usepackage{amsmath,amssymb,amsfonts}
\usepackage{algorithmic}
\usepackage{graphicx}
\usepackage{textcomp}
\usepackage{xcolor}
\def\BibTeX{{\rm B\kern-.05em{\sc i\kern-.025em b}\kern-.08em
    T\kern-.1667em\lower.7ex\hbox{E}\kern-.125emX}}
    
\usepackage{placeins}
\usepackage{graphicx}
\usepackage{subcaption}    
\begin{document}
\newcommand{\squeezeupppp}{\vspace{-8 mm}}
\newcommand{\squeezeuppp}{\vspace{-6 mm}}
\newcommand{\squeezeupp}{\vspace{-5 mm}}
\newcommand{\squeezeup}{\vspace{-3 mm}}
\newcommand{\squeezeu}{\vspace{-2 mm}}
\newcommand{\squeeze}{\vspace{-1 mm}}
\newcommand{\squeez}{\vspace{-.5 mm}}

\title{Multi-task Learning Approach for Automatic Modulation and Wireless Signal Classification}

\author{\IEEEauthorblockN{Anu Jagannath}

\IEEEauthorblockA{Marconi-Rosenblatt AI/ML Innovation Laboratory, \\
ANDRO Computational Solutions, LLC, Rome, NY, USA, \\ E-mail: ajagannath@androcs.com 
}
\and
\IEEEauthorblockN{Jithin Jagannath}
\IEEEauthorblockA{Marconi-Rosenblatt AI/ML Innovation Laboratory, \\
ANDRO Computational Solutions, LLC, Rome, NY, USA,\\ E-mail: jjagannath@androcs.com
}

\thanks{ACKNOWLEDGMENT OF SUPPORT AND DISCLAIMER:(a) Contractor acknowledges Government’s support in the publication of this paper. This material is based upon work supported by the US Army Contract No. W9113M-20-C-0067. (b) Any opinions, findings and conclusions or recommendation expressed in this material are those of the author(s) and do not necessarily reflect the views of the US Army.}
}
\maketitle

\begin{abstract}
Wireless signal recognition is becoming increasingly more significant for spectrum monitoring, spectrum management, and secure communications. Consequently, it will become a key enabler with the emerging fifth-generation (5G) and beyond 5G communications, Internet of Things networks, among others. State-of-the-art studies in wireless signal recognition have only focused on a single task which in many cases is insufficient information for a system to act on. In this work, for the first time in the wireless communication domain, we exploit the potential of deep neural networks in conjunction with multi-task learning (MTL) framework to simultaneously learn modulation and signal classification tasks. The proposed MTL architecture benefits from the mutual relation between the two tasks in improving the classification accuracy as well as the learning efficiency with a lightweight neural network model. Additionally, we consider the problem of heterogeneous wireless signals such as radar and communication signals in the electromagnetic spectrum. Accordingly, we have shown how the proposed MTL model outperforms several state-of-the-art single-task learning classifiers while maintaining a lighter architecture and performing two signal characterization tasks simultaneously. Finally, we also release the only known open heterogeneous wireless signals dataset that comprises of radar and communication signals with multiple labels.
\end{abstract}


\section{Introduction}
Wireless signal recognition plays a vital role in the modern era of wireless communication where heterogeneous wireless entities belonging to civilian, commercial, government, and military applications share the electromagnetic spectrum. Recent years have witnessed an explosive growth of Internet of Things (IoT) devices in critical applications such as smart healthcare, smart industry, smart cities, smart homes, smart vehicles, among others \cite{JagannathAdHoc2019}. The diverse and large scale IoT deployment leads to critical security vulnerabilities in addition to spectrum scarcity. Wireless signal recognition is an emerging technique to identify and mitigate the security weaknesses as well as enable cooperative spectrum sharing to maximize spectrum utility. Signal recognition can be defined as the process of extracting the signal descriptors (modulation, signal type, hardware intrinsic features, etc.,) to characterize the radio frequency (RF) waveform. Spectrum sharing to improve spectrum utilization serves as a key enabler for fifth-generation (5G) and beyond 5G (B5G) communications whereby the various emitters in the vicinity are sensed and identified to allocate and utilize spectrum in a cooperative manner. Realizing the need for improved spectrum sharing to sustain communications, the Federal Communications Commission (FCC) has allocated Citizens Broadband Radio Service (CBRS) in the 3.5 GHz radio band. 
The CBRS band will be cooperatively shared between commercial and government agencies such that only 150 MHz is utilized at a time. This is facilitated by sensing and distinguishing between naval radar and commercial cellular communication systems such that the incumbent naval radar and satellite emissions are not hindered. Additionally, in the tactical front, the wireless signal identification will enhance the spectrum situational awareness allowing soldiers to distinguish between friendly and hostile forces in the battlefield. 

Signal recognition is a widely studied topic, however, it has been segmented into subtasks such as modulation recognition \cite{AlexGoogleNet_AMC,Jagannath18ICC, oshea2018, Jagannath17CCWC, vhf_amc,Jagannath19MLBook, Zhou_AMC_survey}, 
signal type (wireless technology) classification \cite{SignalRecogSurvey}, etc., and studied independently. Furthermore, most of the recent works in this realm focuses either on common communication waveforms \cite{AlexGoogleNet_AMC,Jagannath18ICC,oshea2018,vhf_amc} or radar signals \cite{radar_recog}. 
In a tactical setting as well as in the current scenario of spectrum sharing between government and commercial entities, radar as well as communication waveforms are required to coexist. Therefore, it is essential to consider both categories of waveforms in the signal recognition problem. Additionally, it is important to design a framework that can perform multiple tasks simultaneously to provide more comprehensive information regarding the signal. Consequently, in this work, we propose to jointly model the wireless signal recognition problem in a novel parallel multi-task setting for radar as well as communication waveforms.  


\section{Related Works}

Machine learning is becoming a key enabler for several aspects of wireless communication and radio frequency (RF) signal analysis.  
One of the most common tasks of wireless signal recognition is automatic modulation classification whereby the modulation type of the RF waveform is predicted by the receiver. The modulation classification performance of convolutional neural networks (CNNs) on eight modulation types was studied in \cite{AlexGoogleNet_AMC}. The authors adopted GoogLeNet and AlexNet 
CNN architectures utilizing constellation images as input. However, the employed architectures demonstrated increased reliance on the image preprocessing factors such as image resolution, cropping size, selected area, etc., and achieved an accuracy below 80\% at 0 dB signal-to-noise ratio (SNR). In \cite{Jagannath18ICC}, a feature-based modulation classification with feed-forward neural networks was proposed and demonstrated on USRP software-defined radios with 98\% accuracy for seven modulation classes. Radar waveform recognition on seven classes with a CNN architecture was investigated in \cite{radar_recog}. 
The radar recognition involved feeding time-frequency images to the network in contrast to raw inphase-quadrature (IQ) samples. Single-task modulation classification with CNN on seven classes was studied in \cite{vhf_amc}. The model utilizes cyclic spectrum images as input and was shown to achieve a modulation classification accuracy of 95\% above 2 dB. These approaches use transformed representation or handcrafted features which limit the generalization capability of neural networks in extracting hidden representations from raw IQ signal samples. 

The work by \cite{ICAMCNet} used IQ samples as input to study the performance of a CNN architecture with four convolutional, two pooling, and two dense layers in classifying 11 modulations while achieving an accuracy of 83.4\% at 18 dB. 
A modified ResNet architecture is adopted in \cite{oshea2018} to perform single-task modulation classification on 24 modulation formats. The network achieves a classification accuracy of 95.6\% at 10 dB.  A multi-task learning (MTL) framework for modulation recognition is proposed in \cite{mtlmod} for communication waveforms. They segment a single modulation classification task into subtasks. Hence, their proposed model do not perform multiple tasks simultaneously. These approaches perform a single-task modulation classification on communication waveforms alone. However, our proposed MTL model performs both modulation and signal classification on communication as well as radar waveforms to represent heterogeneous environment. 
In addition, our proposed MTL architecture achieves a modulation classification accuracy of over 99\% above 4 dB on the noise impaired waveforms.

Another subtask of wireless signal recognition is signal classification whereby the wireless technology/standards adopted to generate the RF waveform is accurately identified. Wireless interference detection with a CNN architecture 
were studied by \cite{wirelessInterference}. Three wireless standards namely; IEEE 802.11 b/g, IEEE 802.15.4, and IEEE 802.15.1 occupying different frequency channels were classified into 15 different classes with the highest accuracy attained for IEEE 802.15.1. Wireless technology identification with a CNN architecture to mitigate spectrum crunch in the industrial, scientific, and medical (ISM) band was proposed in \cite{wirelesstech}. Wireless standards such as Zigbee, WiFi, Bluetooth, and their cross-interference representing heterogeneous operation comprising a total of seven classes were classified but required operation in high SNR regime to portray 93\% accuracy. Here again, these works considered single-task signal classification on communication waveforms. In contrast, our work considers both modulation and signal classification tasks on communication and radar waveforms impaired with more dynamic and realistic effects.


Deep learning has made significant strides in the field of computer vision \cite{DL_CV,alexnet}, natural language processing \cite{nlp}, speech recognition \cite{DL_speech}, autonomous control \cite{grigorescu2020survey, Jagannath20UAVBook} etc. 
The comparatively slower pace of applied deep learning in wireless communication in contrast to other domains can be in part attributed to the lack of available large scale datasets for the diverse wireless communication problems.  In this work, we consider a novel MTL model to simultaneously perform two tasks for signal recognition. 
To mitigate the lack of available datasets in the wireless domain and to encourage advances in this area, we release the radar and communication signal dataset developed in this work for open use.

\textbf{Contributions} To the best of our knowledge, our work is the first in the deep learning for wireless communication domain that introduces MTL to solve challenging multiple waveform characterization tasks simultaneously. Unlike the prior works in wireless signal recognition, we propose to jointly model modulation and signal classification as parallel subtasks in an MTL setting. Further, MTL architecture inherently generalizes better with more number of tasks since the model learns shared representation that captures all tasks. Hence, in the future, additional signal classification or regression tasks can be included in the architecture. The novel MTL architecture performs both modulation and signal classifications with over 99\% accuracy above 4 dB on the noise impaired waveforms. We present an elaborate study on the various hyperparameter settings and their effects on the training and classification performances to arrive at a lighter MTL architecture. The proposed MTL architecture is contrasted with several of its single-task learning (STL) counterparts in the literature to depict the MTL advantage in learning parallel tasks with the lighter model. Finally, to motivate future research in this domain, we release the first-of-its-kind radar and communication waveforms dataset with multiple labels for public use \cite{data}.

\section{Wireless Multi-task Learning}

Wireless RF signals can take multiple modulation formats. For example: IEEE802.11a OFDM waveform can possess binary phase-shift keying (BPSK), quadrature phase-shift keying (QPSK), and quadrature amplitude modulation (QAM) modulations. Similarly, satellite communication signals can have M-ary phase-shift keying (PSK) modulations. Several radar signals namely; Airborne-detection, Airborne-range, Air-Ground-MTI, and Ground mapping adopt pulsed continuous wave (PCW) modulation but differ in the transmission parameters such as pulse repetition rate, pulse width, and carrier frequency. Finally, AM radio signals can carry either amplitude modulated double side-band (AM-DSB) or amplitude modulated single side-band (AM-SSB) waveforms. \textit{Hence, it is essential to not merely identify the modulation format but also the signal type to accurately recognize the waveform.}

Multi-task learning (MTL) is a neural network paradigm for inductive knowledge transfer which improves generalization by learning shared representation between related tasks. MTL improves learning efficiency and prediction accuracy on each task in contrast to training an STL model for each task \cite{mtl_95}. MTL has been applied to natural language processing (NLP) and computer vision extensively. 
Unlike NLP and computer vision, MTL has never been applied in the wireless communication realm to the best of our knowledge. In this work, however, we propose to take advantage of the mutual relation between tasks in learning them with an MTL architecture. We adopt a hard parameter shared MTL model \cite{Caruana1993} where the hidden layers among all tasks are shared while preserving certain task-specific layers. Hard parameter sharing significantly reduces the risk of overfitting by the order of the number of tasks as shown by \cite{Baxter1997}. As the model learns more tasks, it extracts shared representation that captures all of the tasks thereby improving the generalization capability of the model. Including additional tasks to the model will, therefore, improve the learning efficiency of the model. Modulation and signal classification are related tasks that can benefit from each other with the hard parameter MTL model. 
Further, such an architecture has the added advantage to benefit from additional tasks motivating the possibility to include future signal characterization tasks. Given an input signal, the proposed MTL model will classify the signal as belonging to a specific modulation and signal class. The modulation and signal classification tasks are optimized with categorical cross-entropy losses denoted by $\mathcal{L}_{m}$ and $\mathcal{L}_{s}$ respectively. The overall multi-task loss ($\mathcal{L}_{mtl}$) function is represented as a weighted sum of losses over the two tasks as in equation (\ref{eq:mtlloss}).
\begin{equation}
\mathcal{L}_{mtl}(\theta_{sh},\theta_{m},\theta_{s})=w_{m}\mathcal{L}_{m}(\theta_{sh},\theta_{m}) + w_{s}\mathcal{L}_{s}(\theta_{sh},\theta_{s})
    \label{eq:mtlloss}
\end{equation}
Here, the joint multi-task loss is parameterized by the shared ($\theta_{sh}$) as well as task-specific ($\theta_{m}, \theta_{s}$) parameters. The weights over the task-specific losses are denoted by $w_{m}$ and $w_{s}$. The MTL training is denoted as the optimization in equation (\ref{eq:mtlopt}).
\begin{align}
    \theta^* = \underset{\theta_{sh},\theta_{m},\theta_{s}}{\arg\min}\mathcal{L}_{mtl}(\theta_{sh},\theta_{m},\theta_{s}) \label{eq:mtlopt}
\end{align}
The MTL optimization aims to tune the network parameters such as to minimize the overall task loss.

\begin{figure}[t]
\centering
\includegraphics[width=.94\columnwidth]{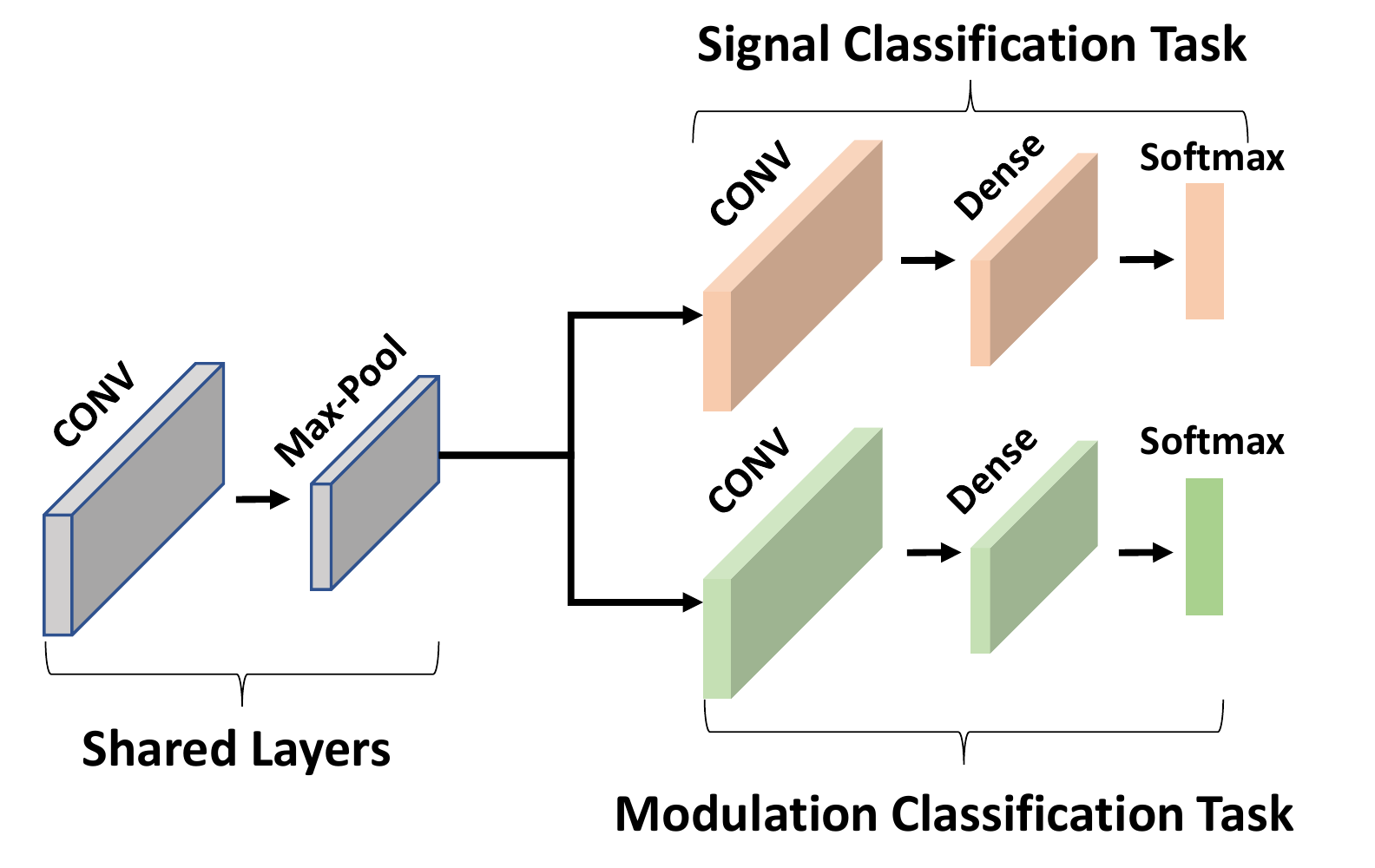} 
\caption{MTL architecture for wireless signal recognition}
\label{fig:mtl}
\end{figure}

\textbf{MTL Network Architecture:} The hard parameter shared MTL architecture for wireless signal recognition is shown in Fig. \ref{fig:mtl}. The shared hidden layers are composed of convolutional and max-pooling layers. Each task-specific branch comprises of convolutional, fully-connected, and output softmax classification layers. The convolutional and fully-connected layers in the network adopt ReLU activation function. 

The hyperparameters such as number of neurons per layer, number of layers, task loss weights, etc., and their effects on the training performance and classification accuracies were studied in-depth as elaborated in the upcoming sections. We train the network with Adam 
gradient descent solver for 30 epochs with a patience of 5. The learning rate is set to 0.001. The architecture adopts batch normalization prior to ReLU activation. The dropout rate of the shared layer is set to 0.25 and that of the task-specific branches are set to 0.25 and 0.5 in the convolutional and fully-connected layers respectively. Unless otherwise stated all the kernel sizes in the convolutional layers are $3\times3$ and the max-pooling size is $2\times2$. The signal and modulation classification task branches perform softmax classification on 11 signal and 9 modulation classes for the noise impaired waveforms (RadComAWGN). We implement our models in Keras with Tensorflow backend on an Ubuntu 18.04 VM running on an Intel Core i5-3230M CPU.

\section{Proposed Model Design and Analysis}

\textbf{Dataset and Evaluation Setting:} As ours is the first work in this realm that proposes an MTL architecture for wireless signal recognition, there are no preexisting datasets that could be leveraged with labels for multiple tasks. Hence, we generate our datasets of radar and communication signals in GNU Radio companion 
for varying SNRs. We generate 2 datasets with modeled propagation effects - RadComAWGN and RadComDynamic at sample rate of 10 MS/s. RadComAWGN comprises a total of 9 modulation and 11 signal classes. The modulation classes are PCW, frequency modulated continuous wave (FMCW), BPSK, AM-DSB, AM-SSB, amplitude shift keying (ASK), Gaussian frequency-shift keying (GFSK), direct sequence spread spectrum complementary code keying (DSSS-CCK), and direct sequence spread spectrum Offset Quadrature Phase-Shift Keying (DSSS-OQPSK). The signal classes are Airborne-detection, Airborne-range, Air-Ground-MTI, Ground mapping, Radar-Altimeter, Satcom, AM Radio, Short-Range, Bluetooth, IEEE802.11bg, and IEEE802.15.4. Of which the first 5 are radar waveforms and the remaining are communication waveforms. The last 3 signal classes are extracted from the interference dataset \cite{crawdad}. Except the last 3, all the waveforms are generated in GNU Radio with additive white Gaussian noise (AWGN) under varying SNR levels (-20 dB to 18 dB in steps of 2 dB). The RadComDynamic dataset contains all waveforms in RadComAWGN except the 3 waveforms from the interference dataset. The waveforms in the RadComDynamic dataset are subject to propagation effects and hardware uncertainties such as multipath, fading, scattering, doppler effects, oscillator drift, and sampling clock offset as shown in Table \ref{tab:dyn}. The propagation channel is chosen to be Rician with K-factor 3. The dataset is partitioned into 70\% training, 20\% validation, and 10\% testing sets. The hyper-parameter evaluations were performed with the RadComAWGN dataset. To benefit future research in MTL on RF signal analysis, we make the dataset publicly available \cite{data}.
\begin{table}[]
\vspace{.3 cm}
\centering
\caption{RadComDynamic: Dynamic settings}
\label{tab:dyn}
\def\arraystretch{1.2}
\begin{tabular}{|p{6.4 cm}|p{1.4cm}|}
\hline
\textbf{Dynamic Parameters}                                   & \textbf{Value}       \\ \hline
Carrier frequency offset std. dev/sample        & 0.05 Hz            \\ \hline
Maximum carrier frequency offset                              & 250 Hz             \\ \hline
Sample rate offset std. dev/sample              & 0.05 Hz            \\ \hline
Maximum sample rate offset                                    & 60 Hz              \\ \hline
Num. of sinusoids in freq. selective fading        & 5                    \\ \hline
Maximum doppler frequency                                     & 2 Hz                 \\ \hline
Rician K-factor                                               & 3                    \\ \hline
Fractional sample delays comprising power delay profile (PDP) &$[0.2, 0.3, 0.1]$ \\ \hline
Number of multipath taps                                      & 5                    \\ \hline
List of magnitudes corresponding to each delay in PDP         & $[1, 0.5, 0.5 ]$   \\ \hline
\end{tabular}
\end{table}
\subsection{Wireless Signal Representation}
Let us denote the generated signal vector as $\mathbf{x}^{id}$ where the superscript $id$ represents the signal key used to extract the signal from the database. The generated signals are complex (IQ) samples of length 128 samples each. The signals are normalized to unit energy prior to storing them in the dataset to remove any residual artifacts from the simulated propagation effects. Data normalization allows a neural network to learn the optimal parameters quickly thereby improving the convergence properties. The normalized data containing both I and Q samples can be denoted as $\hat{\mathbf{x}}^{id} = \hat{\mathbf{x}}^{id}_I + j\hat{\mathbf{x}}^{id}_Q$. Since neural networks can only deal with real numbers, we will vectorize the complex number as below
$\hat{\mathbf{x}}^{id}$
\begin{equation}
   f\{\hat{\mathbf{x}}^{id}\} = \begin{bmatrix} \hat{\mathbf{x}}^{id}_I\\\hat{\mathbf{x}}^{id}_Q\end{bmatrix} \in \mathbb{R}^{256\times1}
\end{equation}
Mathematically, this can be shown with the relation
\begin{equation}
    f:\mathbb{C}^{128\times1} \longrightarrow \mathbb{R}^{256\times1}
\end{equation}
The 256-sample input signal is reshaped to a 2D tensor of size $16\times16$ prior to feeding into the network. The waveforms are stored as key-value pairs in the HDF5 database such that the value can be extracted using the key. The waveform key is denoted by $id={modulation\;format, signal\;class, SNR, sample\;number}$ which matches it to the corresponding waveform in the database.
\subsection{Effect of Task Weights}

In this subsection, we will study the effect of task-specific loss weights on the classification accuracy of both tasks. Specifically, the classifier accuracy on both tasks when the signal strength is very low (SNR$=-2$ dB) will be analyzed. Detection of even the weakest power signal corresponds to improved detection sensitivity. 

\begin{figure}[t!]
\centering
\includegraphics[width=0.85\columnwidth]{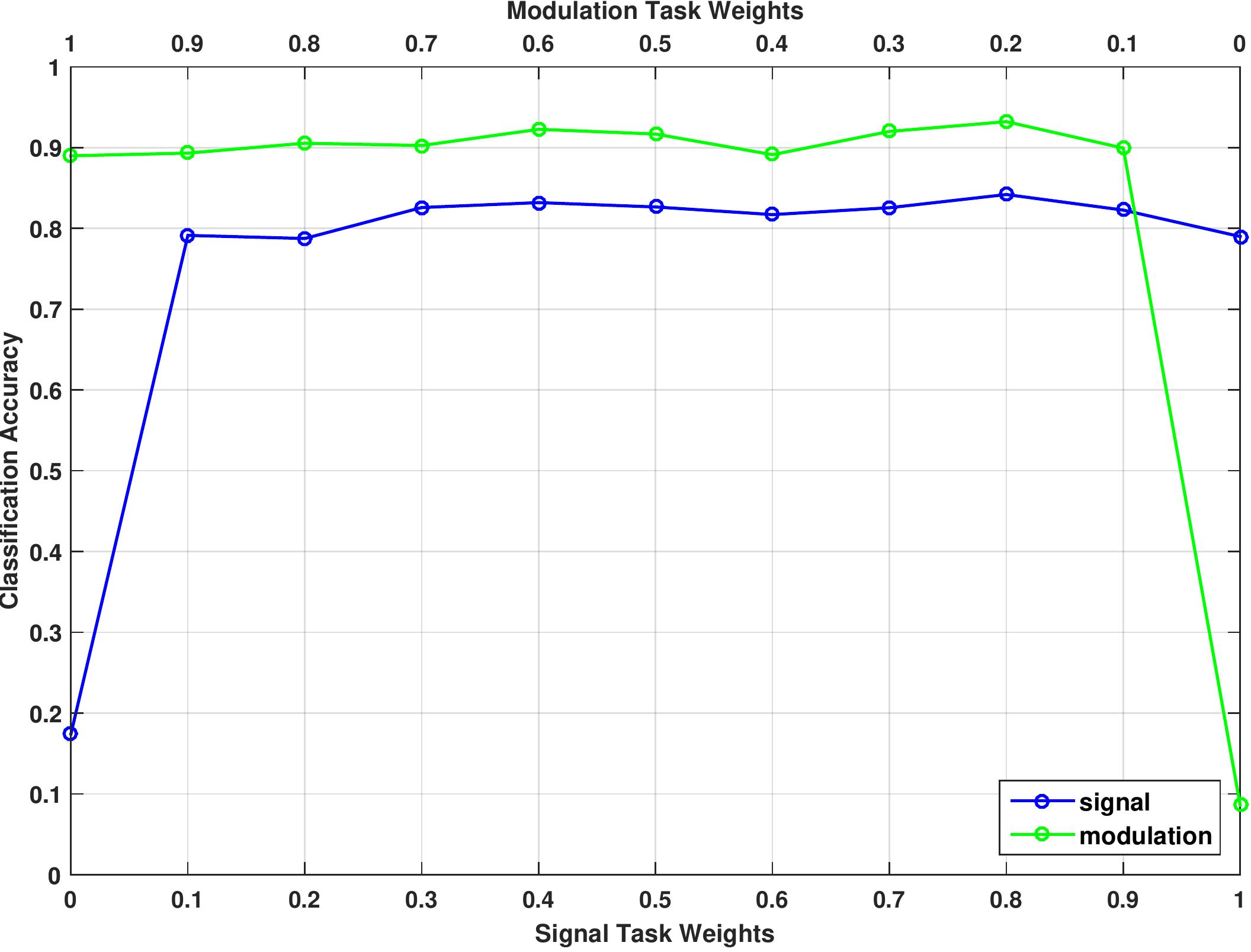} 
\caption{Effect of task loss weight distribution on modulation and signal classification tasks at very low SNR ($-2$ dB)} 
\label{fig:task_weight}  
\end{figure}

Figure \ref{fig:task_weight} shows the classification accuracy of MTL on both tasks at a very low SNR of $-2$ dB for varying weights. The number of kernels in the shared and task-specific convolutional layers are $8$ and $4$ respectively and the number of neurons in the fully-connected layers of the task-specific branches is $256$. The weight distribution for both tasks are varied from $0$ to $1$  in steps of $0.1$ such that sum of weights is unity. The boundaries of the plot denote classification accuracies when the model was trained on individual tasks, i.e., when weights of either task losses were set to zero. It can be seen that the model performs almost stable across the weighting ($0.1$ to $0.9$ on either task). Although for some optimal weighting of $w_{s} = 0.8$ and $w_{m} = 0.2$, both tasks are performing slightly better than at other task weights. We therefore fix the loss weights for both tasks at $w_{s} = 0.8$ and $w_{m} = 0.2$ for the proposed MTL architecture. 

\subsection{Effect of Network Density}

How dense should the network be ? This is the question we are trying to answer in this section. Resource constrained radio platforms require lightweight neural network models for implementation on field programmable gate arrays and application-specific integrated circuits. For such realistic implementations, dense neural network models for signal characterization such as the resource-heavy AlexNet and GoogLeNet adopted by \cite{AlexGoogleNet_AMC} would seem impractical. Hence, rather than adopting dense computer vision models, we handcraft the MTL architecture to arrive at a lighter model. The network density has a direct effect on the learning efficiency and classification accuracy of the model. We will vary the number of neurons in the MTL model introduced in Fig. \ref{fig:mtl} and analyze the effect of introducing additional layers in the shared as well as task-specific branches.

\begin{figure}[h]
\centering
\includegraphics[width=0.85\columnwidth]{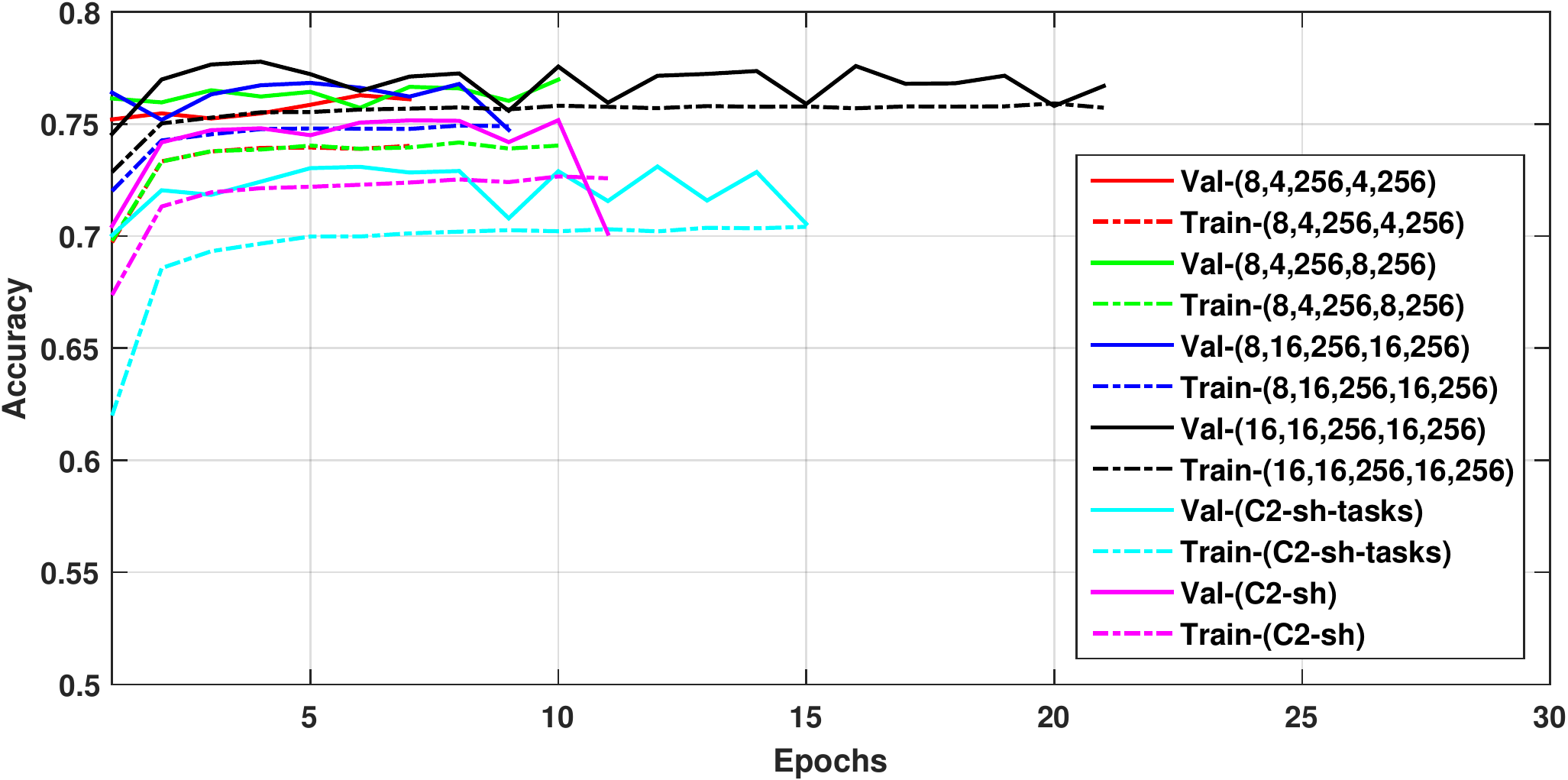} 
\caption{MTL training performance on modulation classification task for varying network density}
\label{fig:mod_nwtrain}   
\end{figure}

\begin{figure}[h]
\centering \vspace{-.2 cm}
\includegraphics[width=0.85\columnwidth]{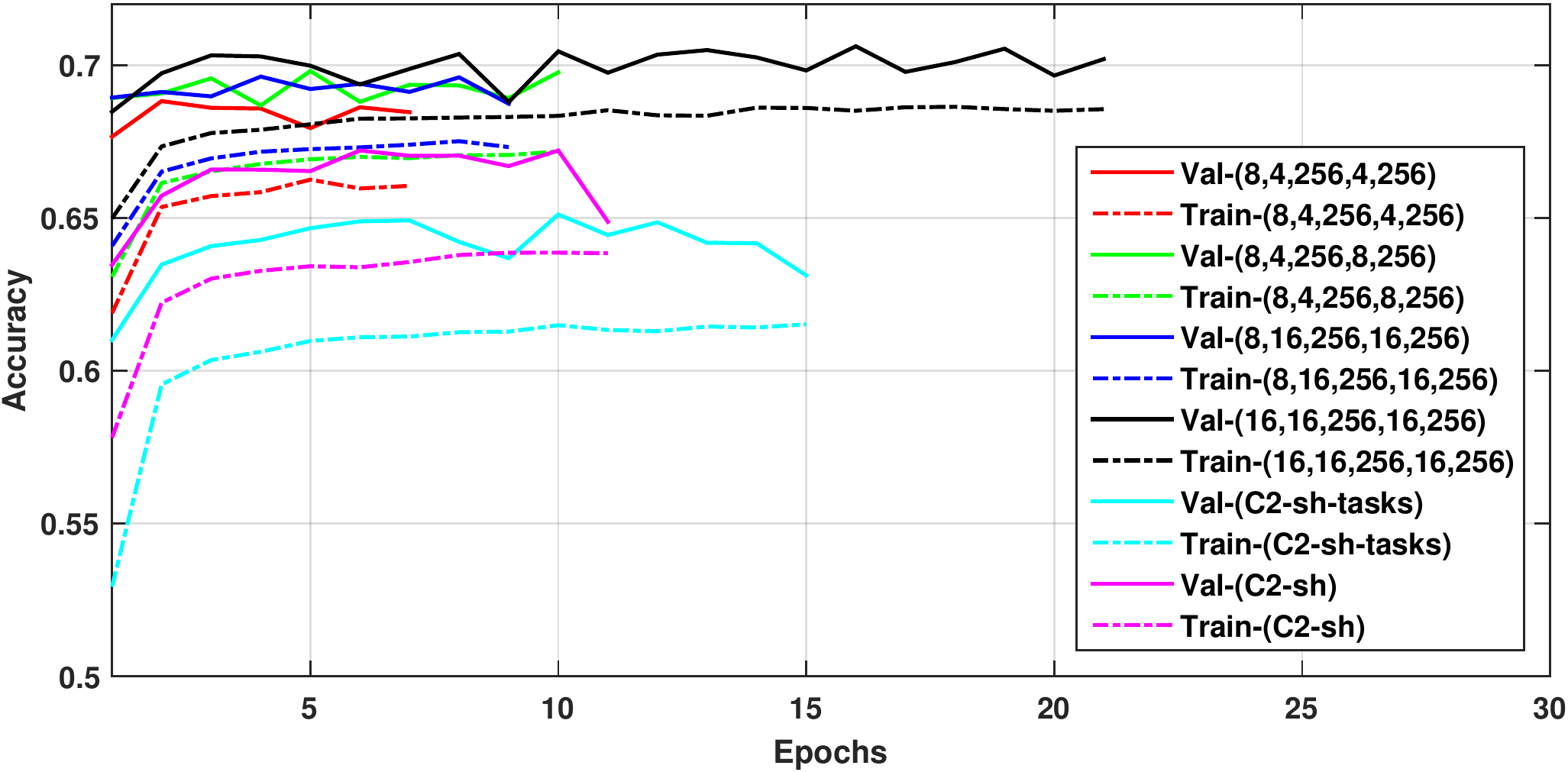} 
\caption{MTL training performance on signal classification task for varying network density}
\label{fig:sig_nwtrain}   
\end{figure}

\begin{figure}[h]
\centering
\includegraphics[width=0.85\columnwidth]{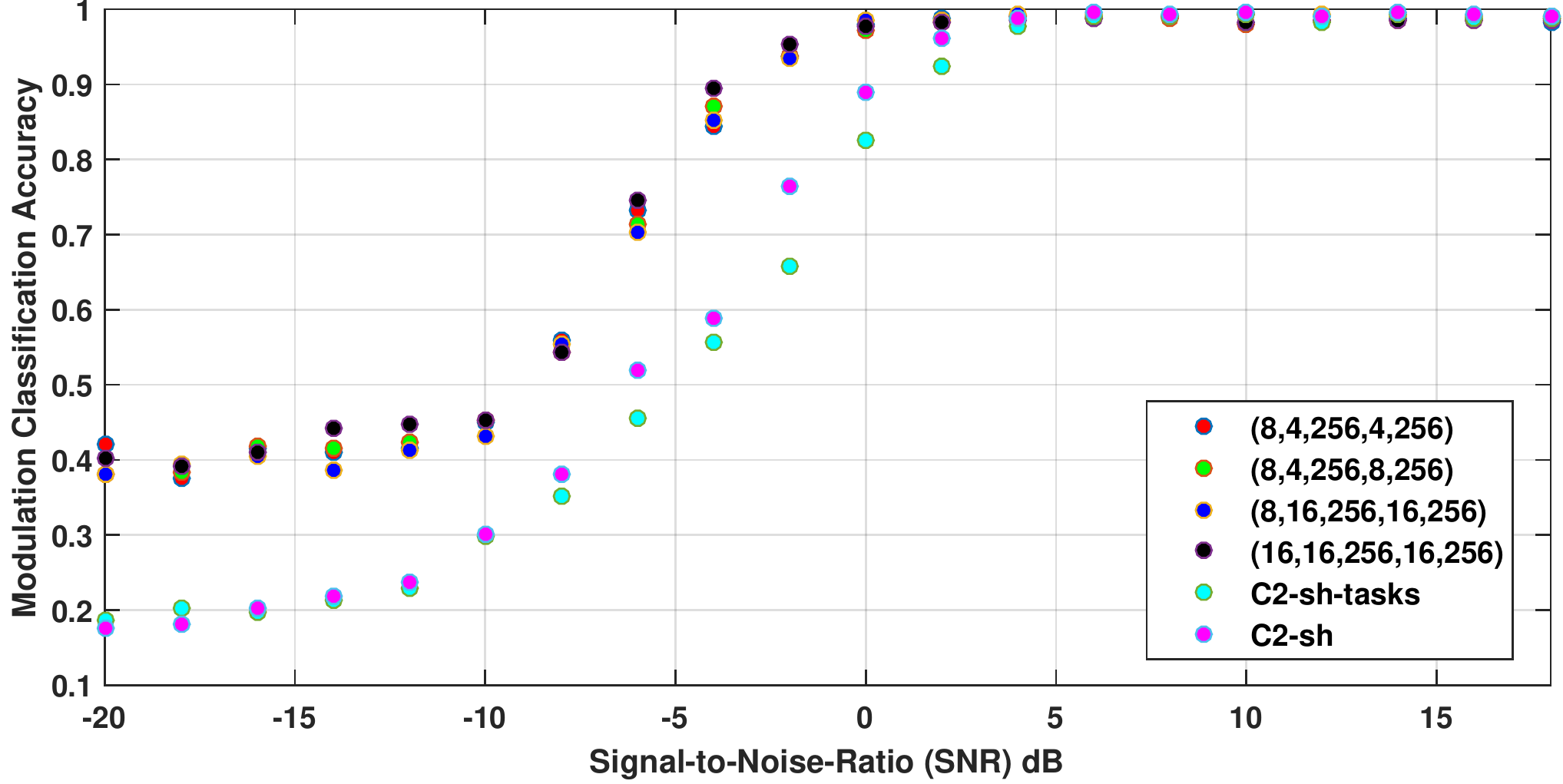} 
\caption{MTL classification performance under varying noise levels on modulation task for varying network density}
\label{fig:mod_nw}  
\end{figure}

\begin{figure}[h!]
\centering \vspace{-.2 cm}
\includegraphics[width=0.84\columnwidth]{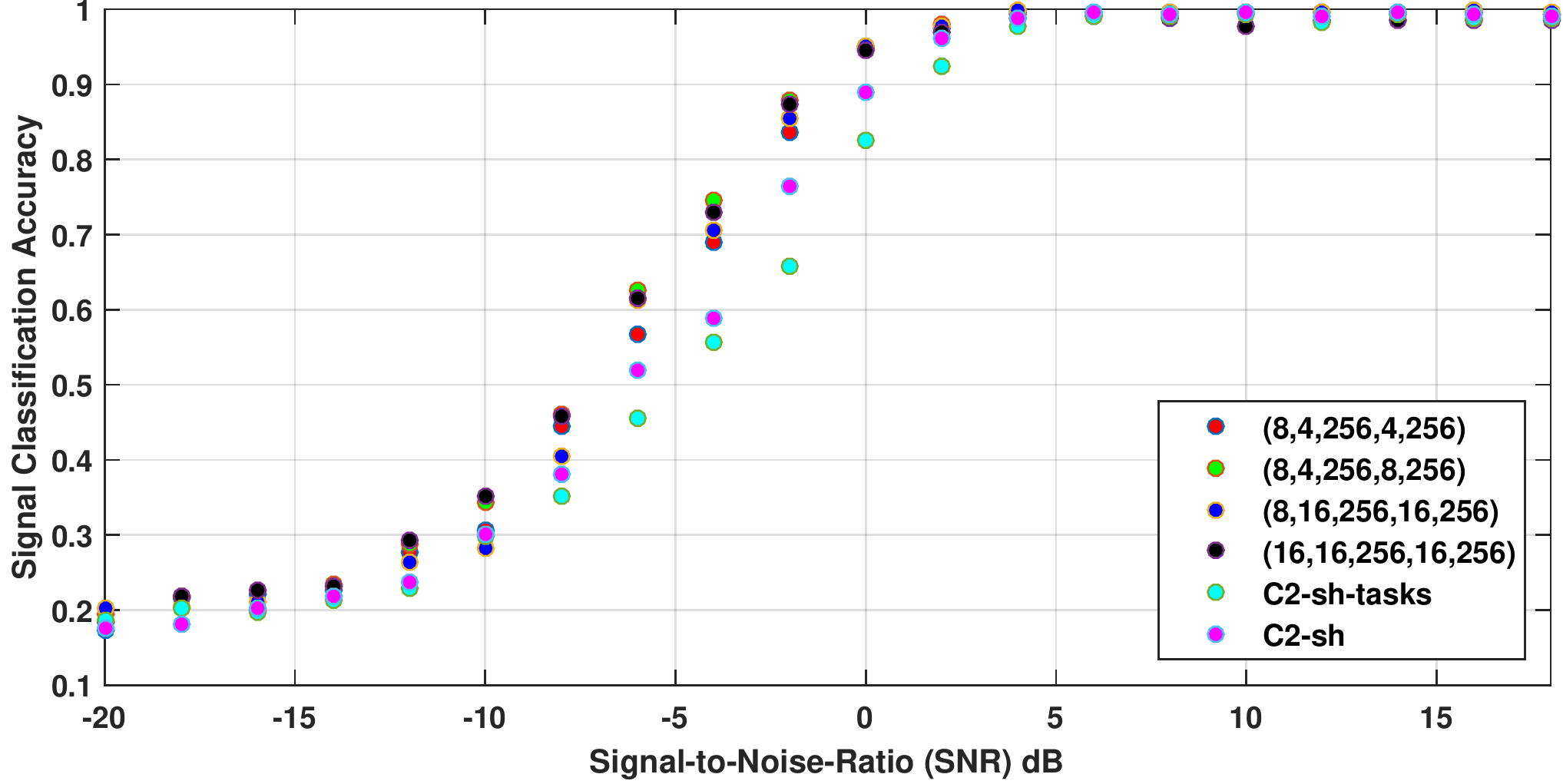} 
\caption{MTL classification performance under varying noise levels on signal task for varying network density}
\label{fig:sig_nw}  
\end{figure}

The legends in the figures (Figure \ref{fig:mod_nwtrain} - Figure \ref{fig:sig_nw}) represent the varying number of neurons as well as layers in the network. The notation $(C_{sh},C_{m},F_{m},C_{s},F_{s})$ implies neuron distribution with $C_{sh},C_{m},C_{s}$ representing the number of filters in the convolutional layer of shared, modulation, and signal branches and $F_{m}, F_{s}$ denote the number of neurons in the fully-connected layers in the modulation and signal branches. The additional layer inclusion notations are $C2-sh$ and $C2-sh-tasks$. The notation $C2-sh$ denotes the MTL architecture with two convolutional layers each followed by a max-pooling layer in the shared module. The number of filters in the convolutional layers of the shared module is 8. Finally, $C2-sh-tasks$ denote the MTL model with shared module architecture the same as $C2-sh$ but with two sequential convolutional layers in the task-specific branches. The number of filters in the convolutional layers of both task-specific branches is 4. The number of neurons in the fully-connected layers of task-specific branches is 256 for both $C2-sh$ and $C2-sh-tasks$.

Figure \ref{fig:mod_nwtrain} and Figure \ref{fig:sig_nwtrain} show the training performance of the MTL model with respect to the two tasks. The training plots demonstrate that increasing the network density slows the training speed of the model. This is intuitive as the network parameters increase training time increases. The fastest network training time is achieved with the model configuration of $(8,4,256,4,256)$ which is the lightest of all configurations. Figure \ref{fig:mod_nw} and Figure \ref{fig:sig_nw} demonstrate the classification accuracies on both tasks for varying network density under increasing SNR levels (decreasing noise power). It can be seen that the additional layers in the shared ($C2-sh$) and shared as well as task-specific branches ($C2-sh-tasks$) does not improve the classification accuracies but rather results in significantly poor modulation and signal classification accuracies. Further, the MTL model does not seem to benefit from the remaining dense configurations. Hence, the MTL model will use the lighter configuration of $(8,4,256,4,256)$ that yields better learning efficiency and prediction accuracies.

\begin{table*}[!h]
\caption{Comparison of proposed MTL with other STL models}
\centering
\def\arraystretch{1.6}%
\begin{tabular}{|p{3.4cm}|p{3.4cm}|p{3.2cm}|p{2.4cm}|p{2.4cm}|}
\hline
\textbf{Model}       & \textbf{Modulation Classification  Accuracy}       & \textbf{Signal Classification \newline Accuracy}       &\textbf{Number of Classes}       &\textbf{Waveform Type}       \\ \hline
\multicolumn{5}{|c|}{\textbf{Modulation and Signal classification (this work) - Multi-task}} \\ \hline
Proposed MTL Model         &97.87\% at 0 dB, \newline 99.53\% at 10 dB                &92.3\% at 0 dB,\newline 99.53\% at 10 dB                &9 modulation,\newline 11 signal classes              &Radar and \newline Communication                   \\ \hline
\multicolumn{5}{|c|}{\textbf{Modulation classification only methods - Single Task}}                                  \\ \hline
Peng et al. 2019 \cite{AlexGoogleNet_AMC}        & below 80\% at 0 dB               & -              & 8              &Communication                    \\ \hline
Jagannath et al. 2018 \cite{Jagannath18ICC}        & 98\% above ~25 dB               & -              & 7              &Communication                    \\ \hline
O'Shea et al. 2018 \cite{oshea2018}        & 95.6\% at 10 dB               & -              & 24              &Communication                    \\ \hline
Mossad et al. 2019 \cite{mtlmod}        & 86.97\% at 18 dB               & -              & 10              &Communication                    \\ \hline
Hermawan et al. 2020 \cite{ICAMCNet}        &$\sim$80\% at 0 dB, \newline 83.4\% at 18 dB              & -              & 11              &Communication                    \\ \hline
Wang et al. 2017 \cite{radar_recog}        &100\% at 0 dB              & -              & 7              &Radar                    \\ \hline
Li et al. 2018 \cite{vhf_amc}        &95\% above 2 dB              & -              & 7              &Communication    \\ \hline
\multicolumn{5}{|c|}{\textbf{Signal classification only methods - Single Task}}                                   \\ \hline
Bitar et al. 2017 \cite{wirelesstech}        &-              &91\% at 15-25 dB, \newline 93\% at 30 dB              & 7              &Communication    \\ \hline
Schmidt et al. 2017 \cite{wirelessInterference}  &-              &95\% at -5 dB             & 15              &Communication    \\ \hline
\end{tabular}
\\
\label{tabl}
\end{table*}

\begin{figure*}[!ht]
        \centering
        \begin{subfigure}[b]{0.32\textwidth}
            \centering
\includegraphics[width=0.99\columnwidth]{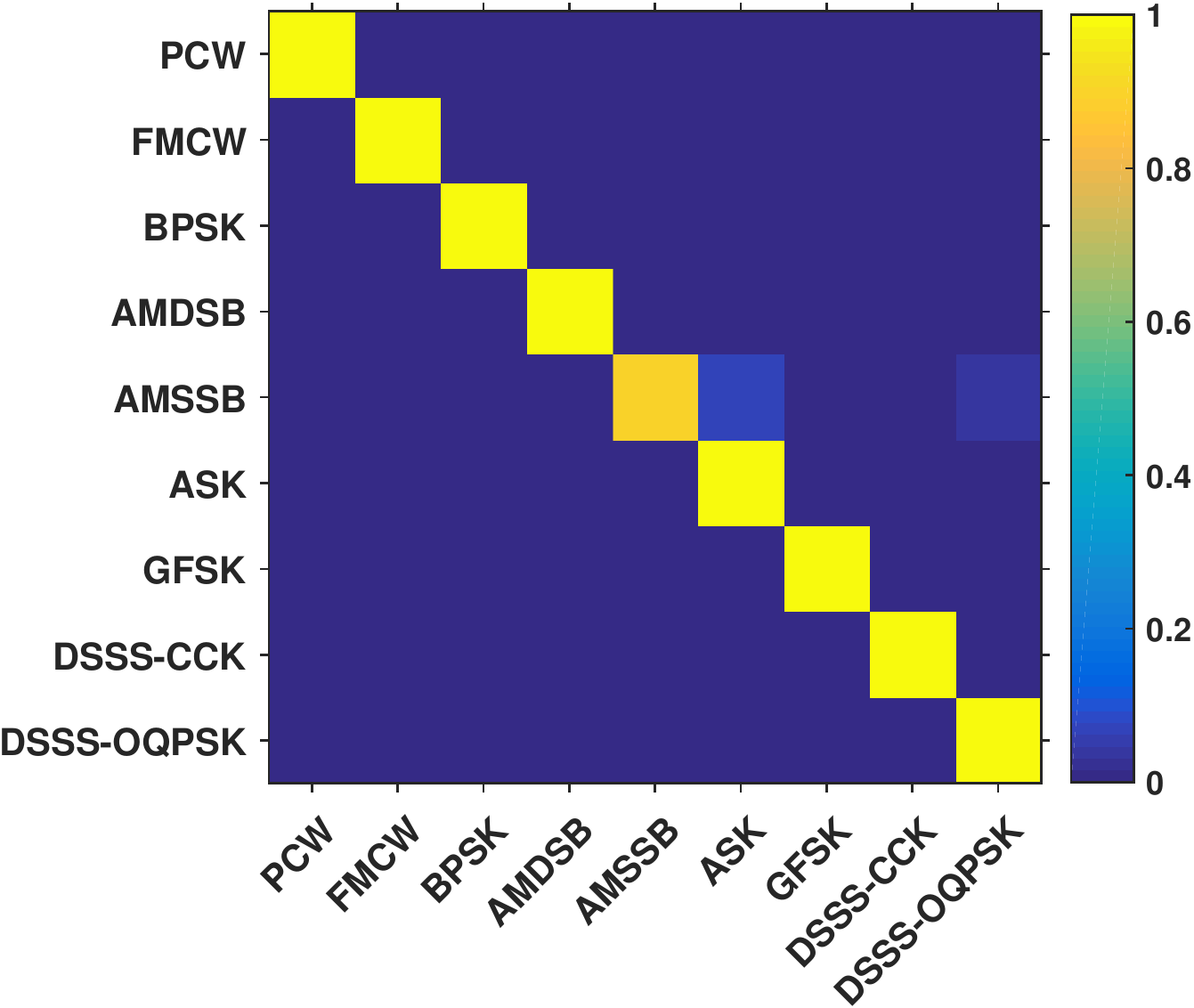} 
\caption{RadComAWGN: Mod. classification}
\label{fig:cm_radcomawgn_mod}    
        \end{subfigure}\hspace{2mm}
        \begin{subfigure}[b]{0.30\textwidth}   
           \centering
\includegraphics[width=0.99\columnwidth]{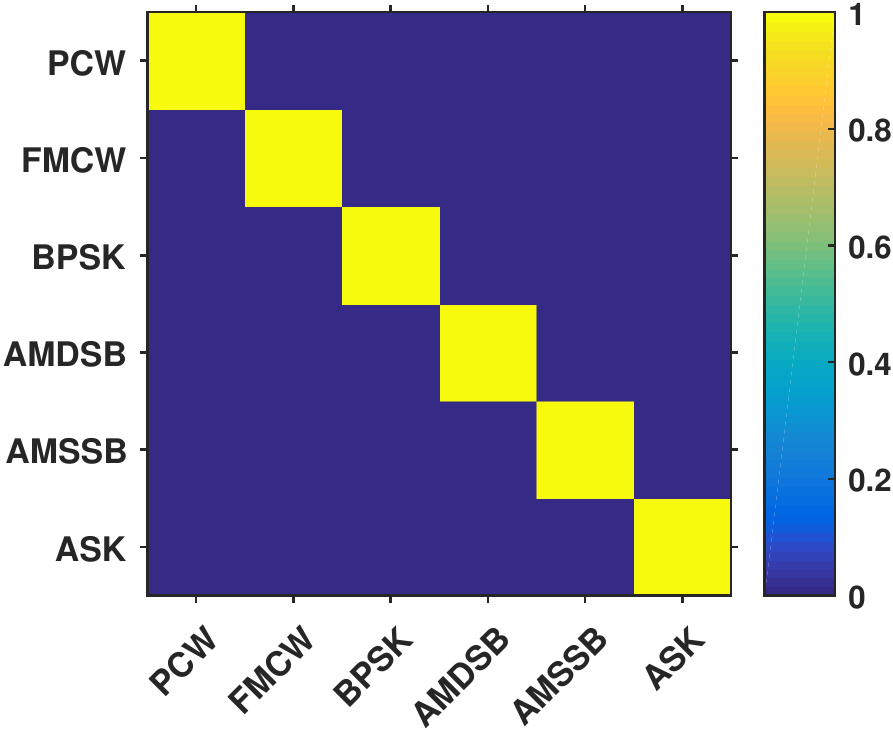}\vspace{4mm} 
\caption{RadComDynamic: Mod. classification}
        \end{subfigure}\hspace{2mm}
        \begin{subfigure}[b]{0.345\textwidth}   
            \centering
\includegraphics[width=0.99\columnwidth]{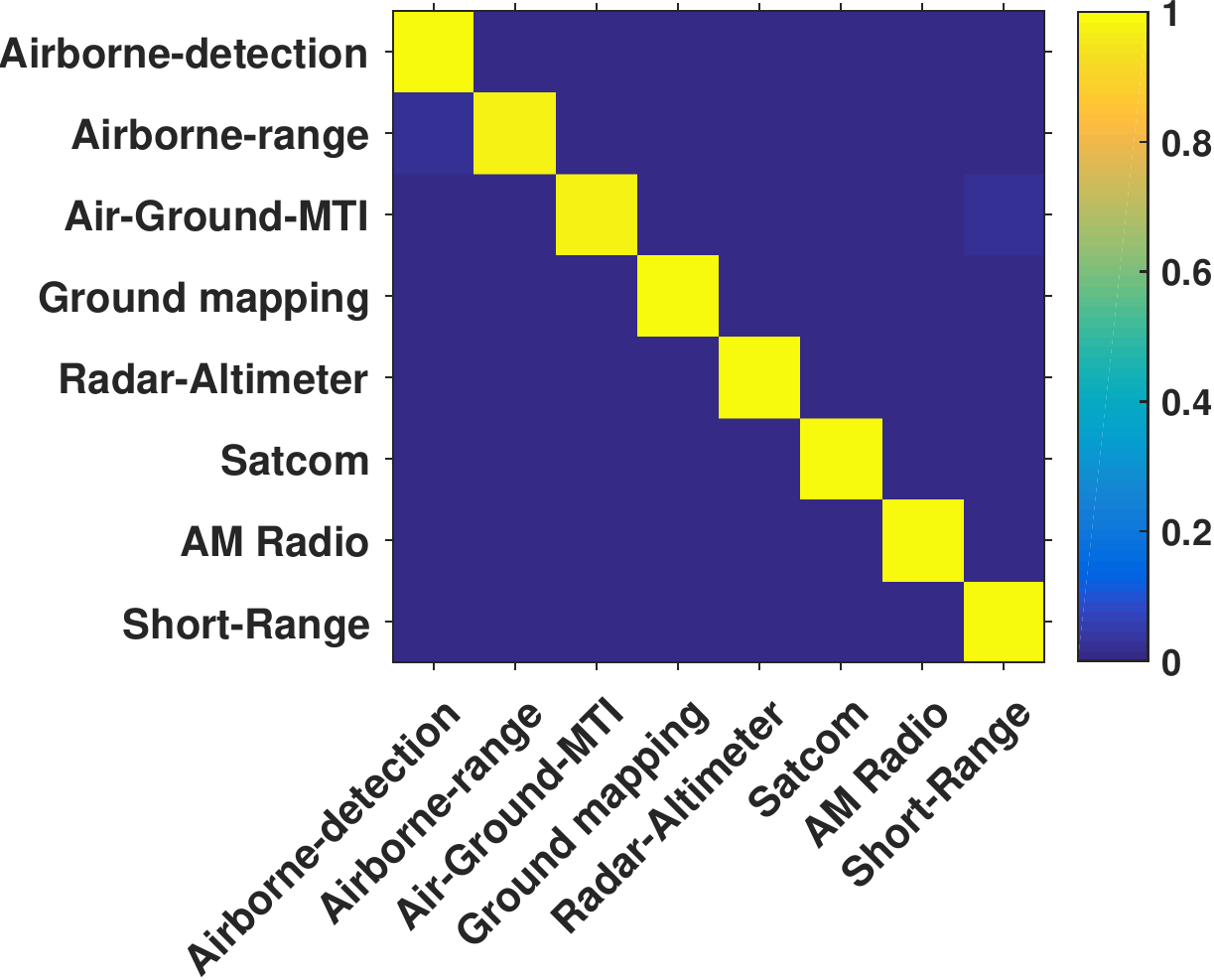}
\caption{RadComDynamic: Signal classification}
        \end{subfigure}\vspace{2mm}
        
        \begin{subfigure}[b]{0.43\textwidth}  
            \centering
\includegraphics[width=0.96\columnwidth]{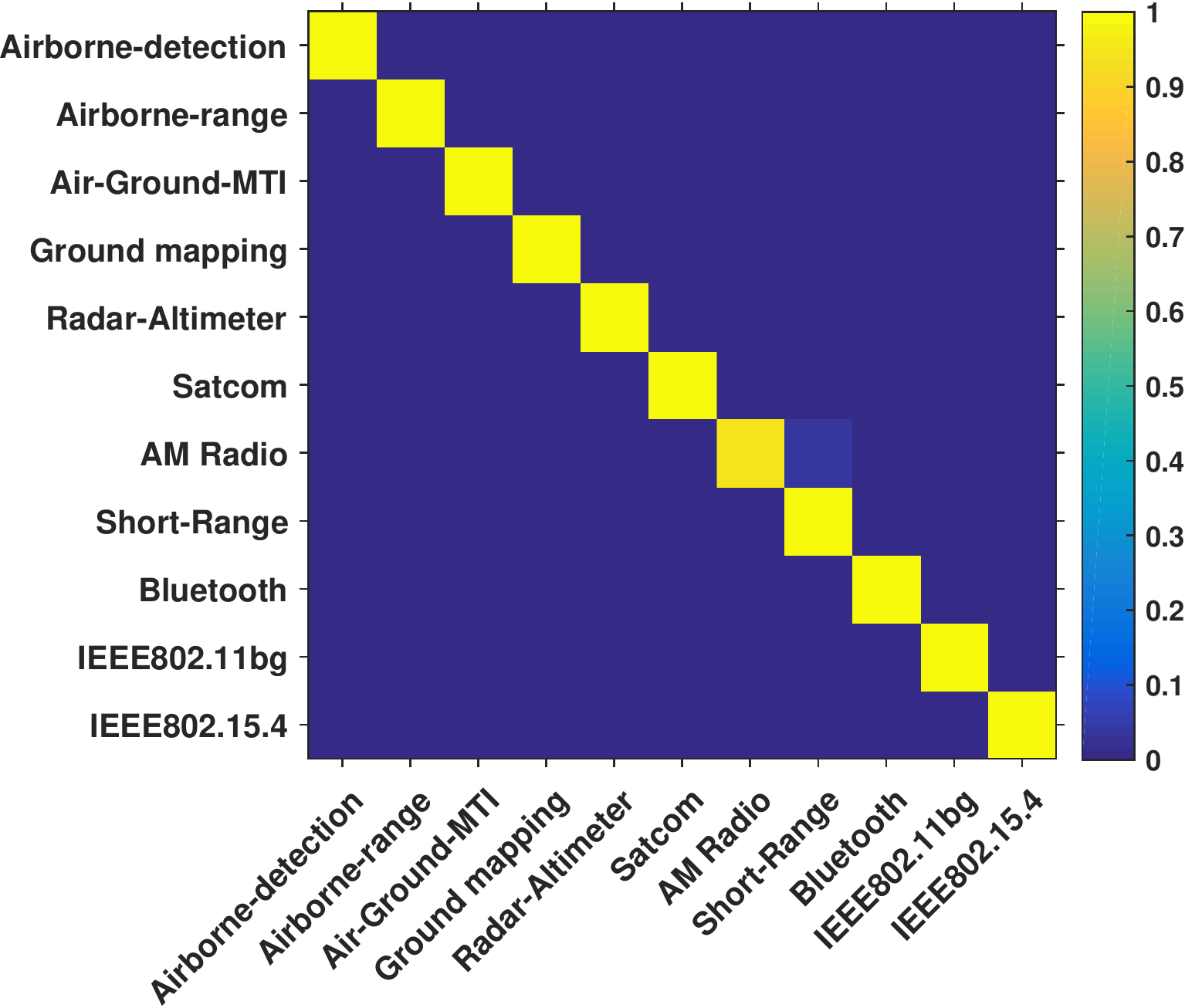} 
\caption{RadComAWGN: Signal classification}
\label{fig:cm_radcomawgn_sig}
        \end{subfigure}\hspace{1mm}
        \begin{subfigure}[b]{0.55\textwidth}   
        \centering
\includegraphics[width=0.97\columnwidth]{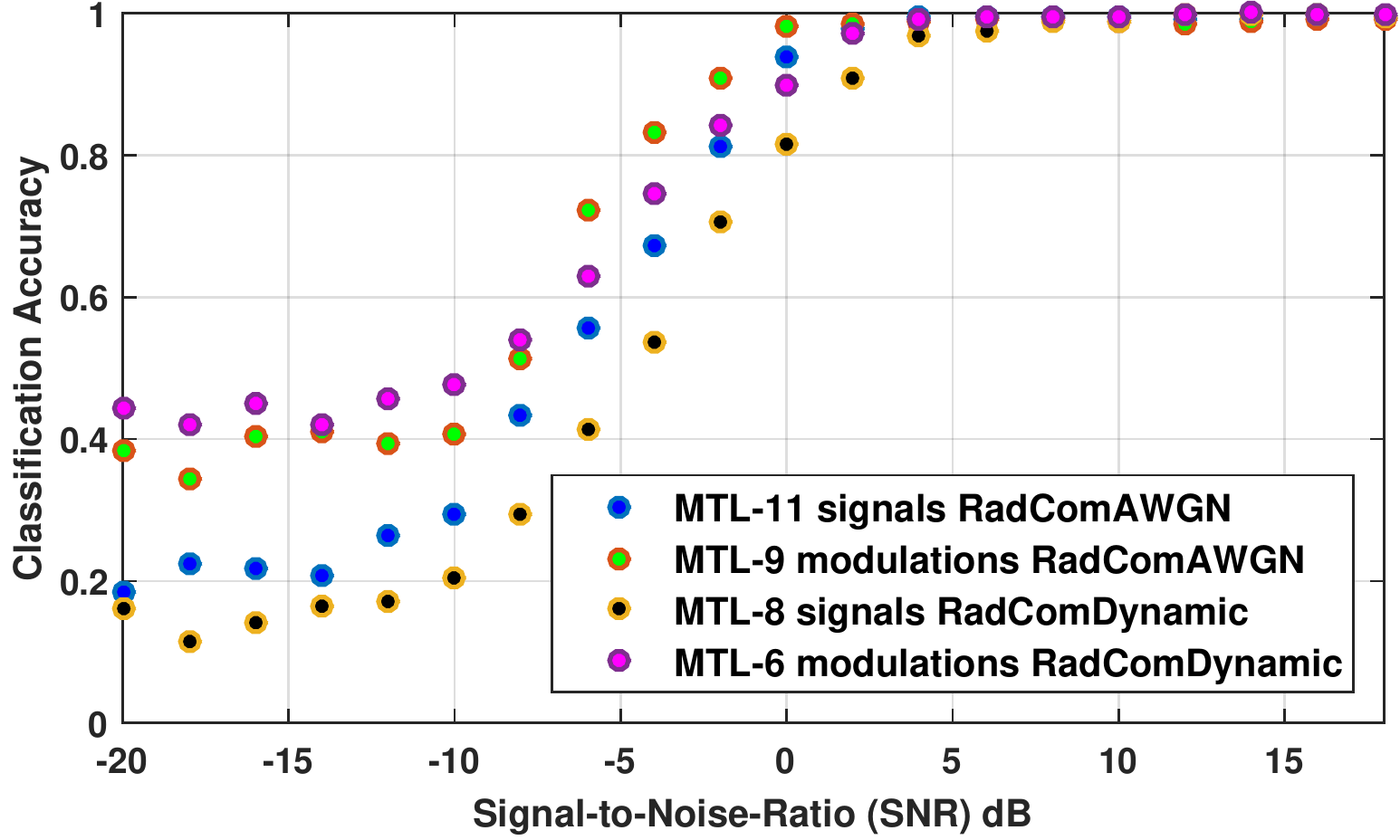} 
\caption{Fine-tuned MTL classification performance under varying SNR}
\label{fig:mtl_acc}
        \end{subfigure}
        \caption
        {\small Fine-tuned MTL confusion matrices on both tasks at SNR= 10 dB (a-d) \& Performance under varying SNR (e)} 
        \label{fig:cm}
    \end{figure*}


\section{Fine-tuned model performance Evaluation}
In this section, we demonstrate the performance of the fine-tuned MTL model on RadComAWGN and RadComDynamic datasets for varying noise levels. With these tests, we are aiming to assess the MTL performance on waveforms impaired by just AWGN as well as waveforms affected by \textit{realistic propagation and radio hardware impairments} (previously discussed in Table \ref{tab:dyn}). We adopted transfer learning on the RadComDynamic experiments by initializing the weights of the network to the tuned weights of MTL with RadComAWGN while the weights were randomly initialized for RadComAWGN tests. The MTL exhibits a 98.58\% modulation classification accuracy on RadComAWGN and 97.07\% on RadComDynamic dataset at 2 dB. The signal classification accuracy of MTL at 2 dB yielded 97.87\% and 90.86\% on RadComAWGN and RadComDynamic datasets respectively. We show that the proposed MTL model yields above 90\% accuracy at SNRs above 2 dB for both tasks with noise impaired (RadComAWGN) as well as propagation and hardware impaired (RadComDynamic) waveforms. The confusion matrices of the signal and modulation classes at 10 dB on RadComAWGN and RadComDynamic datasets along with their classification accuracy at varying noise levels are shown in Fig. \ref{fig:cm}. These experiments demonstrate the classification capability of the proposed lightweight MTL model on severely impaired waveforms under varying signal powers.

In Table \ref{tabl}, we compare the proposed MTL model with other state-of-the-art methods in both the tasks. The classification accuracies of the proposed MTL model are with the RadComAWGN noise impaired waveforms. Our framework is the first method that accomplishes both tasks with a single model. \emph{It is to be noted that in the current literature, to best of our knowledge, there does not exist an MTL model or a multi-task labelled dataset for modulation and signal recognition.} We would require either of these to perform a one-to-one comparison. Nonetheless, we provide a tabular comparison to show the proposed MTL model achieves the same or better accuracy as compared to state-of-the-art STL models. \textit{This proves the utility and effectiveness of using a single MTL model in the RF domain.} 
The single-task modulation classifier proposed in \cite{radar_recog} which achieves a 100\% accuracy at 0 dB is with fewer classes and utilizes handcrafted input features which limits the generalization capability. In contrast, our model adopts a significantly lighter CNN model to achieve two simultaneous tasks on more number of classes.  Additionally, raw IQ samples in our model allow capturing hidden representations improving generalization capability. 
Overall, the proposed lightweight model has provided reliable performance over several varying scenarios outperforming most state-of-the-art STL models. 


\section{Conclusion and Future Work}
We proposed a multi-task learning framework to solve two challenging and fundamental wireless signal recognition tasks - modulation and signal classification. We leveraged the relation between the two tasks in allowing the MTL to learn the shared representation. The classification accuracy and learning efficiency of the two tasks were experimentally demonstrated with the novel lightweight MTL architecture motivating its adoption in resource-constrained embedded radio platforms. The performance of the model was depicted for noise impaired as well as propagation and hardware impaired waveforms. To benefit future research utilizing MTL for wireless communication, we publicly release our dataset. The success of the proposed MTL architecture further opens the door to include more signal characterization tasks such as bandwidth regression, sampling rate regression, pulse width regression, emitter classification, etc., to the model. The inclusion of additional signal characterization tasks will be part of our future research along with generating more waveforms to be included to the dataset for training multi-task frameworks. 


\bibliographystyle{ieeetr}
\bibliography{bibfile}

\begin{thebibliography}{10}

\bibitem{JagannathAdHoc2019}
J.~Jagannath, N.~Polosky, A.~Jagannath, F.~Restuccia, and T.~Melodia, ``Machine
  learning for wireless communications in the internet of things: A
  comprehensive survey,'' {\em Ad Hoc Networks (Elsevier)}, vol.~93, p.~101913,
  2019.

\bibitem{AlexGoogleNet_AMC}
S.~{Peng}, H.~{Jiang}, H.~{Wang}, H.~{Alwageed}, Y.~{Zhou}, M.~M. {Sebdani},
  and Y.~{Yao}, ``Modulation classification based on signal constellation
  diagrams and deep learning,'' {\em IEEE Transactions on Neural Networks and
  Learning Systems}, vol.~30, no.~3, pp.~718--727, 2019.

\bibitem{Jagannath18ICC}
J.~Jagannath, N.~Polosky, D.~O. Connor, L.~Theagarajan, B.~Sheaffer, S.~Foulke,
  and P.~Varshney, ``{Artificial Neural Network based Automatic Modulation
  Classifier for Software Defined Radios},'' in {\em Proc. of IEEE Intl, Conf.
  on Communications (ICC)}, (Kansas City, USA), May 2018.

\bibitem{oshea2018}
T.~J. {O’Shea}, T.~{Roy}, and T.~C. {Clancy}, ``Over-the-air deep learning
  based radio signal classification,'' {\em IEEE Journal of Selected Topics in
  Signal Processing}, vol.~12, no.~1, pp.~168--179, 2018.

\bibitem{Jagannath17CCWC}
J.~Jagannath, D.~O'Connor, N.~Polosky, B.~Sheaffer, L.~N. Theagarajan,
  S.~Foulke, P.~K. Varshney, and S.~P. Reichhart, ``{Design and Evaluation of
  Hierarchical Hybrid Automatic Modulation Classifier using Software Defined
  Radios},'' in {\em Proc. of IEEE Annual Computing and Communication Workshop
  and Conference (CCWC)}, (Las Vegas, NV, USA), January 2017.

\bibitem{vhf_amc}
R.~{Li}, L.~{Li}, S.~{Yang}, and S.~{Li}, ``Robust automated vhf modulation
  recognition based on deep convolutional neural networks,'' {\em IEEE
  Communications Letters}, vol.~22, no.~5, pp.~946--949, 2018.

\bibitem{Jagannath19MLBook}
J.~Jagannath, N.~Polosky, A.~Jagannath, F.~Restuccia, and T.~Melodia, ``{Neural
  Networks for Signal Intelligence: Theory and Practice,},'' in {\em Machine
  Learning for Future Wireless Communications} (F.~Luo, ed.), Wiley - IEEE
  Series, John Wiley \& Sons, Limited, 2020.

\bibitem{Zhou_AMC_survey}
R.~{Zhou}, F.~{Liu}, and C.~W. {Gravelle}, ``Deep learning for modulation
  recognition: A survey with a demonstration,'' {\em IEEE Access}, vol.~8,
  pp.~67366--67376, 2020.

\bibitem{SignalRecogSurvey}
X.~Li, F.~Dong, S.~Zhang, and W.~Guo, ``A survey on deep learning techniques in
  wireless signal recognition,'' {\em Wireless Comms. and Mobile Computing},
  vol.~2019, pp.~1--12, 02 2019.

\bibitem{radar_recog}
C.~{Wang}, J.~{Wang}, and X.~{Zhang}, ``Automatic radar waveform recognition
  based on time-frequency analysis and convolutional neural network,'' in {\em
  Proc. of IEEE International Conference on Acoustics, Speech and Signal
  Processing (ICASSP)}, pp.~2437--2441, 2017.

\bibitem{ICAMCNet}
A.~P. {Hermawan}, R.~R. {Ginanjar}, D.~{Kim}, and J.~{Lee}, ``Cnn-based
  automatic modulation classification for beyond 5g communications,'' {\em IEEE
  Communications Letters}, vol.~24, no.~5, pp.~1038--1041, 2020.

\bibitem{mtlmod}
O.~S. {Mossad}, M.~{ElNainay}, and M.~{Torki}, ``Deep convolutional neural
  network with multi-task learning scheme for modulations recognition,'' in
  {\em Proc. of International Wireless Communications Mobile Computing
  Conference (IWCMC)}, pp.~1644--1649, 2019.

\bibitem{wirelessInterference}
M.~{Schmidt}, D.~{Block}, and U.~{Meier}, ``Wireless interference
  identification with convolutional neural networks,'' in {\em Proc. of the
  IEEE Intl. Conf. on Industrial Informatics (INDIN)}, pp.~180--185, 2017.

\bibitem{wirelesstech}
N.~{Bitar}, S.~{Muhammad}, and H.~H. {Refai}, ``Wireless technology
  identification using deep convolutional neural networks,'' in {\em Proc. of
  Intl Symp. on Personal, Indoor, and Mobile Radio Comms. (PIMRC)}, pp.~1--6,
  2017.

\bibitem{DL_CV}
K.~He, X.~Zhang, S.~Ren, and J.~Sun, ``Deep residual learning for image
  recognition,'' {\em Proc. of IEEE Conference on Computer Vision and Pattern
  Recognition (CVPR)}, pp.~770--778, 2016.

\bibitem{alexnet}
A.~Krizhevsky, I.~Sutskever, and G.~E. Hinton, ``Imagenet classification with
  deep convolutional neural networks,'' in {\em Proc. of the 25th International
  Conference on Neural Information Processing Systems - Volume 1}, NIPS’12,
  (Red Hook, NY, USA), p.~1097–1105, 2012.

\bibitem{nlp}
R.~Collobert and J.~Weston, ``A unified architecture for natural language
  processing: Deep neural networks with multitask learning,'' in {\em Proc. of
  the International Conf. on Machine Learning}, ICML ’08, (New York, NY,
  USA), p.~160–167, Association for Computing Machinery, 2008.

\bibitem{DL_speech}
T.~{Afouras}, J.~S. {Chung}, A.~{Senior}, O.~{Vinyals}, and A.~{Zisserman},
  ``Deep audio-visual speech recognition,'' {\em IEEE Transaction on Pattern
  Analysis and Machine Intelligence}, pp.~1--1, 2018.

\bibitem{grigorescu2020survey}
S.~Grigorescu, B.~Trasnea, T.~Cocias, and G.~Macesanu, ``A survey of deep
  learning techniques for autonomous driving,'' {\em Journal of Field
  Robotics}, vol.~37, no.~3, pp.~362--386, 2020.

\bibitem{Jagannath20UAVBook}
J.~Jagannath, A.~Jagannath, S.~Furman, and T.~Gwin, ``Deep learning and
  reinforcement learning for autonomous unmanned aerial systems: Roadmap for
  theory to deployment,'' {\em arXiv preprint 2009.03349}, 2020.

\bibitem{data}
{A. Jagannath and J. Jagannath}, ``Communication and radar dataset for
  modulation and signal classification.''
  {https://github.com/ANDROComputationalSolutions/RadarCommDataset}, 2020.

\bibitem{mtl_95}
S.~Thrun, ``Is learning the n-th thing any easier than learning the first?,''
  in {\em Proc. of the Intl. Conf. on Neural Information Processing Systems},
  NIPS’95, (Cambridge, MA, USA), p.~640–646, MIT Press, 1995.

\bibitem{Caruana1993}
R.~Caruana, ``Multitask learning: A knowledge-based source of inductive bias,''
  in {\em Proc. of the Intl. Conf. on Machine Learning}, 1993.

\bibitem{Baxter1997}
J.~Baxter, ``A bayesian/information theoretic model of learning to learn
  viamultiple task sampling,'' {\em Mach. Learn.}, vol.~28, p.~7–39, July
  1997.

\bibitem{crawdad}
M.~Schmidt, D.~Block, and U.~Meier, ``{CRAWDAD} dataset owl/interference (v.
  2019-02-12).'' Downloaded from
  {https://crawdad.org/owl/interference/20190212}, Feb. 2019.

\end{thebibliography}

\end{document}